\documentclass[letterpaper, 10 pt, conference]{ieeeconf}

\IEEEoverridecommandlockouts
\usepackage{cite}
\usepackage{amsmath,amssymb}
\usepackage{booktabs}
\usepackage{graphicx}
\usepackage{xcolor}
\usepackage{url}
\usepackage[hidelinks]{hyperref}  % hidelinks: no red/blue boxes around links

\graphicspath{{figures/}}
\title{\LARGE \bf
Grasp-Taxonomy Conditioning as a Mode Switch\\for Dexterous Grasping of Climbing Holds}
\ifdefined\ANONYMOUS
  \author{Anonymous Authors}
  \hypersetup{pdfauthor={}}
\else
  \author{Rumi Loghmani, Abraham George, Amir Barati Farimani%
    \thanks{The authors are with the Department of Mechanical Engineering,
    Carnegie Mellon University.}}
  \hypersetup{pdfauthor={Rumi Loghmani, Abraham George, Amir Barati Farimani}}
\fi

\begin{document}
\maketitle
\thispagestyle{empty}
\pagestyle{empty}

\begin{abstract}
Dexterous grasping in robotics relies on distinct hand configurations depending on
object geometry. We introduce \textbf{BetaGrasp}, a 3D diffusion policy that conditions
a DP3-style point-cloud policy on a four-class grasp label (crimp, sloper, pinch, and
jug), running on a Franka arm with a 16-DoF LEAP hand. We also introduce a
climbing-hold benchmark: a spring testbed whose ratchet gives a slip-force reading in
newtons. In our analysis the policy's point-cloud encoder did not separate the four
holds, so the scan alone gave it little to choose a grasp from. Without the label (a
DP3-style baseline), simulated rollouts from the robot's reset pose all end in a jug
grasp, whatever the hold (25\% correct grasp type, versus 94\% for BetaGrasp). The label
works as a switch for which grasp the policy performs. On 80 paired hardware trials,
BetaGrasp raises average slip force from 9.6\,N to 14.3\,N, and grasps slipping fully off
the hold drop from 44\% of trials to 10\%.
\end{abstract}

% Keywords are entered on the PaperPlaza submission form (first two from the RA-L list):
% Dexterous manipulation; imitation learning; deep learning in grasping and manipulation;
% grasping.

% -----------------------------------------------------------------------
\section{Introduction}
% -----------------------------------------------------------------------

\begin{figure*}[t]
\centering
\includegraphics[width=\textwidth]{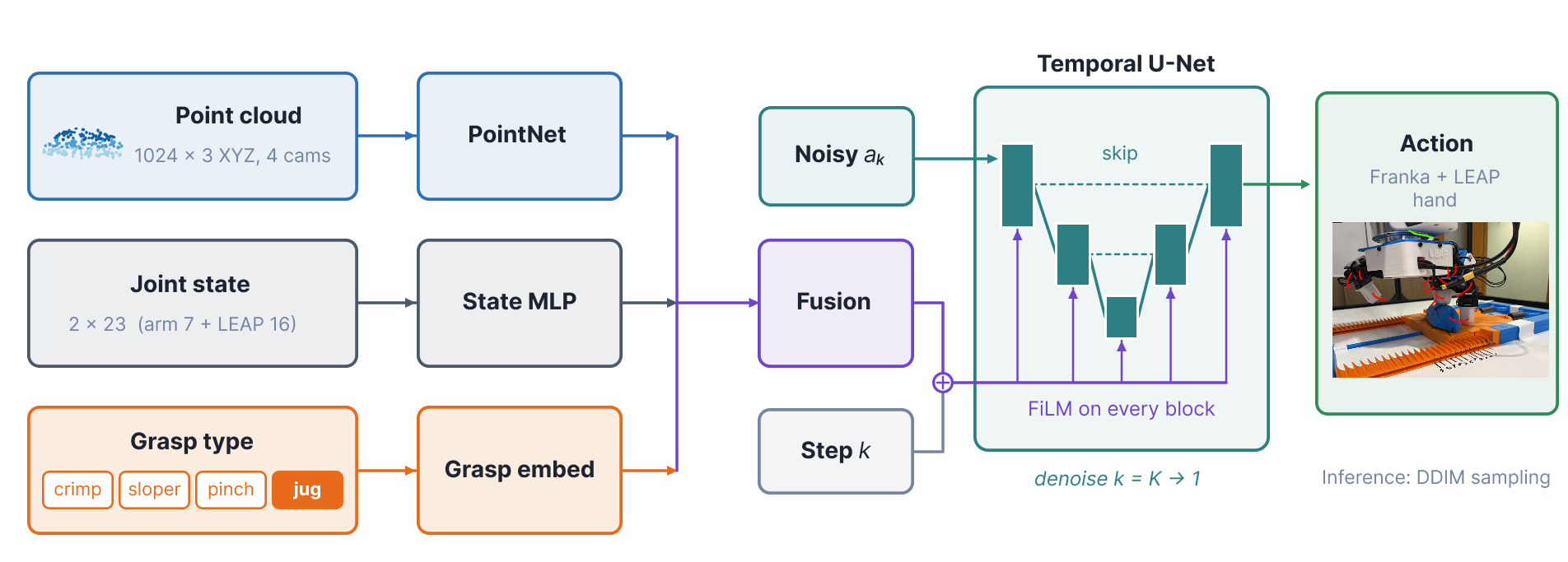}
\caption{BetaGrasp architecture. Point cloud, proprioception, and grasp-type
label are encoded independently and fused into a 512-d conditioning vector $z$
that feature-wise conditions a 1D temporal U-Net diffusion denoiser. The DP3 baseline drops only the grasp-type branch (orange)
(37{,}248 parameters, $0.1\%$ of 34.6\,M).}
\label{fig:arch}
\end{figure*}

Imitation learning based manipulation has developed rapidly in the domain of parallel-jaw claw
gripper end-effectors, while development on multi-finger end effectors has been slower
to follow, because small changes in object geometry call for small changes in how the
fingers are placed. When grasping a climbing hold, the grip used directly determines how much
force the hand can support before slipping off \cite{schweizer2001crimp,amca2012grip}. In
this paper we split one high-level task (``grasp the hold'') into four grasps:
\emph{crimp} (fingertips curled tight on a thin edge), \emph{sloper} (an open hand
pressing on a rounded surface and relying on friction), \emph{pinch} (the thumb
squeezing against the fingers), and \emph{jug} (fingers hooked into a deep handle). Each
grasp uses a different hand shape and set of contacts, and each fails differently under
load.

Diffusion-based behavior cloning is now a common choice for visuomotor policy learning
\cite{chi2025diffusion,ho2020ddpm}. We build on DP3 \cite{ze20243d}, a 3D diffusion
policy that takes point clouds as input (Fig.~\ref{fig:arch}). Most diffusion policies
model grasping as one action distribution conditioned on observations, with nothing in
the input that says which grasp to use. A multi-grasp policy then has to infer the grasp
from the observation alone, which is hard on climbing holds: in our analysis, after
cropping and downsampling to 1024 points, the policy's point-cloud encoder did not
separate the four holds (\S\ref{sec:mechanistic}), even though each needs a different
grasp. Recent taxonomy-conditioned grasping methods
\cite{dexonomy2025,tian2024graspasyousay} do use grasp type, but they generate static
grasp poses, not a policy that executes the grasp in closed loop.

We expected taxonomy conditioning to help as a \emph{mode selector}. Without it, a
policy trained on several grasp types could collapse to one of them, say a jug grasp
every time, and fail on the other three holds. Our simulated rollouts support this.
Starting from the robot's reset pose, the baseline ends in a jug grasp on every hold.

\paragraph*{Contributions}
\begin{enumerate}
  \item A \textbf{grasp-taxonomy-aware 3D diffusion policy}
    (BetaGrasp\footnote{In climbing, ``beta'' is the knowledge of how to do a move.}),
    built on DP3 \cite{ze20243d}, that takes a grasp-type label as input along with the
    point cloud and robot state.
  \item A \textbf{climbing-hold manipulation benchmark}: four holds that each need a
    different grasp, a spring testbed that constrains the hold to one axis, and a linear
    ratchet that records slip force, tested at five hold orientations.
  \item An \textbf{analysis of what the label changes}: in simulated rollouts from the
    reset pose, the policy without it ends in a jug grasp on every hold, and on hardware
    that policy does best on the jug hold and worse on the other three.
\end{enumerate}

% -----------------------------------------------------------------------
\section{Related Work}
% -----------------------------------------------------------------------
\subsection{Diffusion Policies for Robot Control}
Diffusion Policy \cite{chi2025diffusion} treats behavior cloning as conditional
denoising: a temporal U-Net conditioned on observations generates action trajectories.
Diffusion lets it represent multimodal action distributions \cite{ho2020ddpm}, and DDIM
\cite{song2021ddim} shortens inference.

DP3 \cite{ze20243d} replaces image inputs with point clouds, encoded by a lightweight
PointNet-style encoder \cite{qi2017pointnet}. iDP3 \cite{ze2024idp3} adapts it to
humanoid robots, using egocentric point clouds so that no camera calibration is needed.
GenDP \cite{wang2024gendp} adds 3D semantic fields to generalize across object instances
within a category. Diffusion models have also been used to generate dexterous grasps
\cite{weng2024dexdiffuser,wang2023dexgraspnet,xu2023unidexgrasp}.

\subsection{Dexterous Manipulation Data and Embodiments}
DexCap \cite{wang2024dexcap} collects dexterous demonstrations with a portable
motion-capture system. Among teleoperation systems
\cite{qin2023anyteleop,cheng2024opentelevision,zhao2023aloha}, we build directly on the
open-source OpenVR codebase \cite{george2023openvr}, a VR teleoperation system for the
Franka arm originally paired with an Allegro hand. We retarget its hand interface to the
LEAP Hand v1 \cite{shaw2023leap}, chosen for its wider finger range of motion.
UniDexFPM \cite{wu2024unidexfpm} learns functional pre-grasp manipulation with a
diffusion policy distilled from reinforcement-learning teachers, and CrossDex
\cite{crossdex2025} transfers grasping across different hands through an eigengrasp
action space. We instead keep one hand and vary the grasp.

\subsection{Grasp Taxonomy and Type-Conditioned Grasping}
Human grasp taxonomies sort grasps into categories by opposition type, thumb position,
and power versus precision \cite{napier1956prehensile,cutkosky1989grasp,feix2016grasp}.
The closest work is Dexonomy \cite{dexonomy2025}, which synthesizes dexterous grasps for
31 of the 33 types in the GRASP taxonomy \cite{feix2016grasp} and trains a
type-conditioned generator that takes single-view point clouds (82.3\% real-world success on a
Shadow Hand). Its output is a set of static hand poses (pre-grasp, grasp, squeeze), not
a learned closed-loop policy. Language-conditioned grasp generation
\cite{tian2024graspasyousay} also produces static poses. OmniDexVLG
\cite{omnidexvlg2024} uses a vision-language model to reason about grasp type and
affordance, then generates grasp poses from a partial point cloud. DexGraspVLA
\cite{dexgraspvla2026} pairs a vision-language planner with a diffusion execution
policy, but the planner chooses \emph{which object} to grasp, not \emph{how}, and the
policy is not conditioned on grasp type.

We are not aware of prior work that conditions a closed-loop imitation-learning policy
for an arm--hand system on grasp type. Table~\ref{tab:novelty} compares the methods.

% -----------------------------------------------------------------------
\section{Method: Grasp-Taxonomy-Aware 3D Diffusion Policy}
\label{sec:method}
% -----------------------------------------------------------------------

\subsection{Problem Formulation}
Each episode is a grasp of one climbing hold seated on the spring testbed
(\S\ref{sec:testbed}). At each control step $t$, the policy observes (i) a fused point cloud
$P \in \mathbb{R}^{N \times 3}$ (captured once per episode after parking the arm out
of the cameras' field of view), (ii) a proprioceptive state history $s_{t-T_o+1:t}$
over observation horizon $T_o$, and (iii) a grasp-type label $g \in \{0,1,2,3\}$
corresponding to \{crimp, sloper, pinch, jug\}. The policy outputs a joint-space
action chunk $a_{t:t+T_p-1}$ over prediction horizon $T_p$; the controller executes
the first $T_a \le T_p$ steps before replanning. The operator supplies $g$ during data
collection and at deployment.

Following Chi et al.~\cite{chi2025diffusion}, we model the conditional
distribution over action sequences as a denoising diffusion process,
$p(a_{t:t+T_p-1} \mid P, s_{t-T_o+1:t}, g)$, learned from teleoperated demonstrations
via supervised diffusion loss.

% Table I is defined here (not in Related Work) so it floats to the top of the right
% column on page 2; defined earlier, it left a gap at the bottom of the left column.
\begin{table}[t]
\centering\small
\setlength{\tabcolsep}{4pt}
\caption{Positioning against related dexterous-hand methods.}
\label{tab:novelty}
\begin{tabular}{lccc}
\toprule
Method & \shortstack{Point-cloud\\input} & \shortstack{Grasp-type\\conditioning}
  & \shortstack{Trajectory\\policy} \\
\midrule
DP3 \cite{ze20243d}                           & \checkmark & $\times$   & \checkmark \\
Dexonomy \cite{dexonomy2025}                  & \checkmark & \checkmark & $\times$   \\
Grasp as You Say \cite{tian2024graspasyousay} & \checkmark & \checkmark & $\times$   \\
OmniDexVLG \cite{omnidexvlg2024}              & \checkmark & \checkmark & $\times$   \\
DexGraspVLA \cite{dexgraspvla2026}            & $\times$   & $\times$   & \checkmark \\
\textbf{BetaGrasp}                            & \checkmark & \checkmark & \checkmark \\
\bottomrule
\end{tabular}
\end{table}

% Fig. 2 is defined here (on page 2) because a full-width figure* can only appear at
% the top of a LATER page; defined in Sec. IV-B it landed on page 4.
% Panels share one height so they align regardless of each photo's aspect ratio.
\begin{figure*}[t]
\centering
\setlength{\tabcolsep}{2pt}
\begin{tabular}{ccc}
\includegraphics[height=4.2cm]{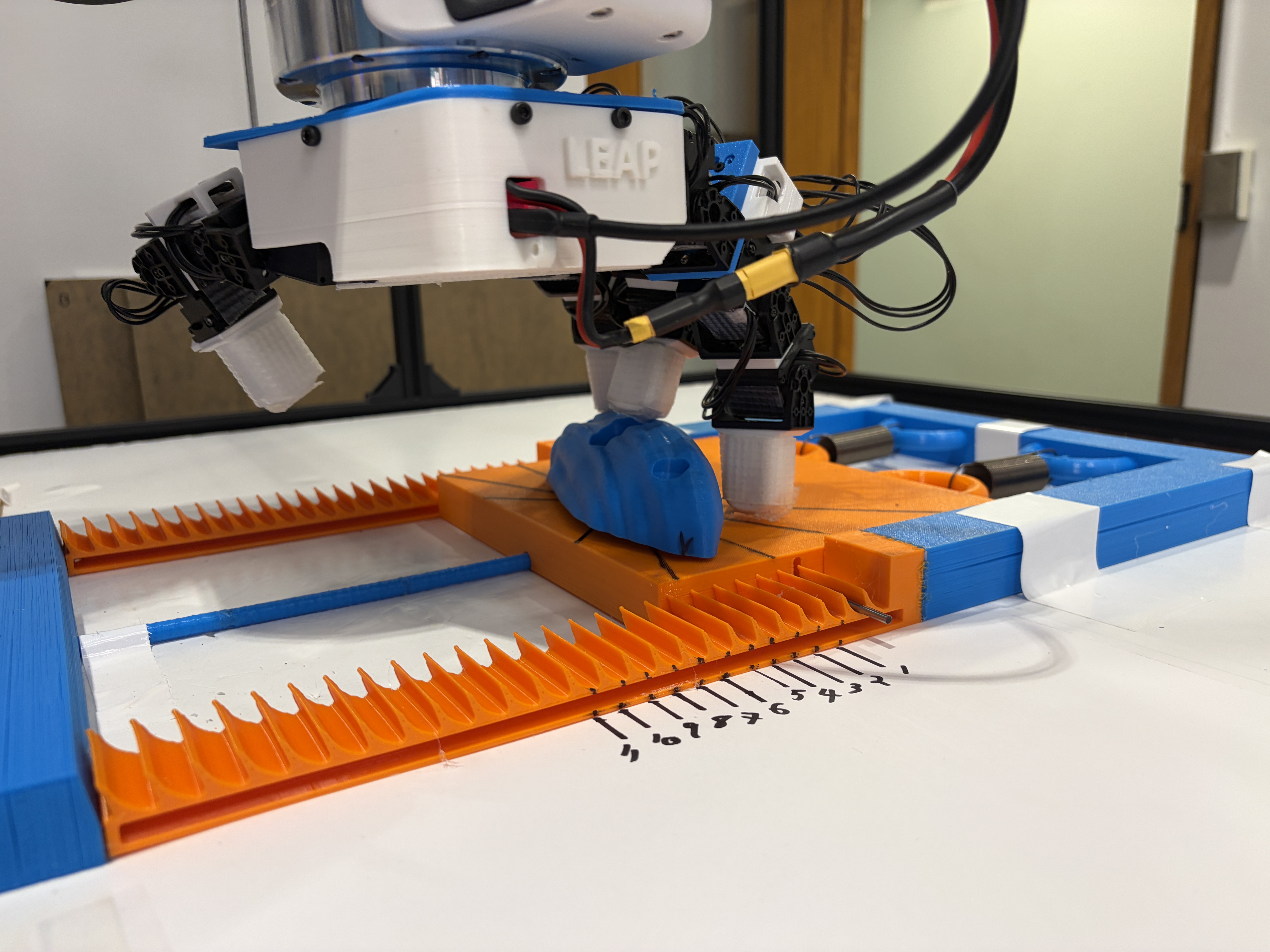} &
\includegraphics[height=4.2cm]{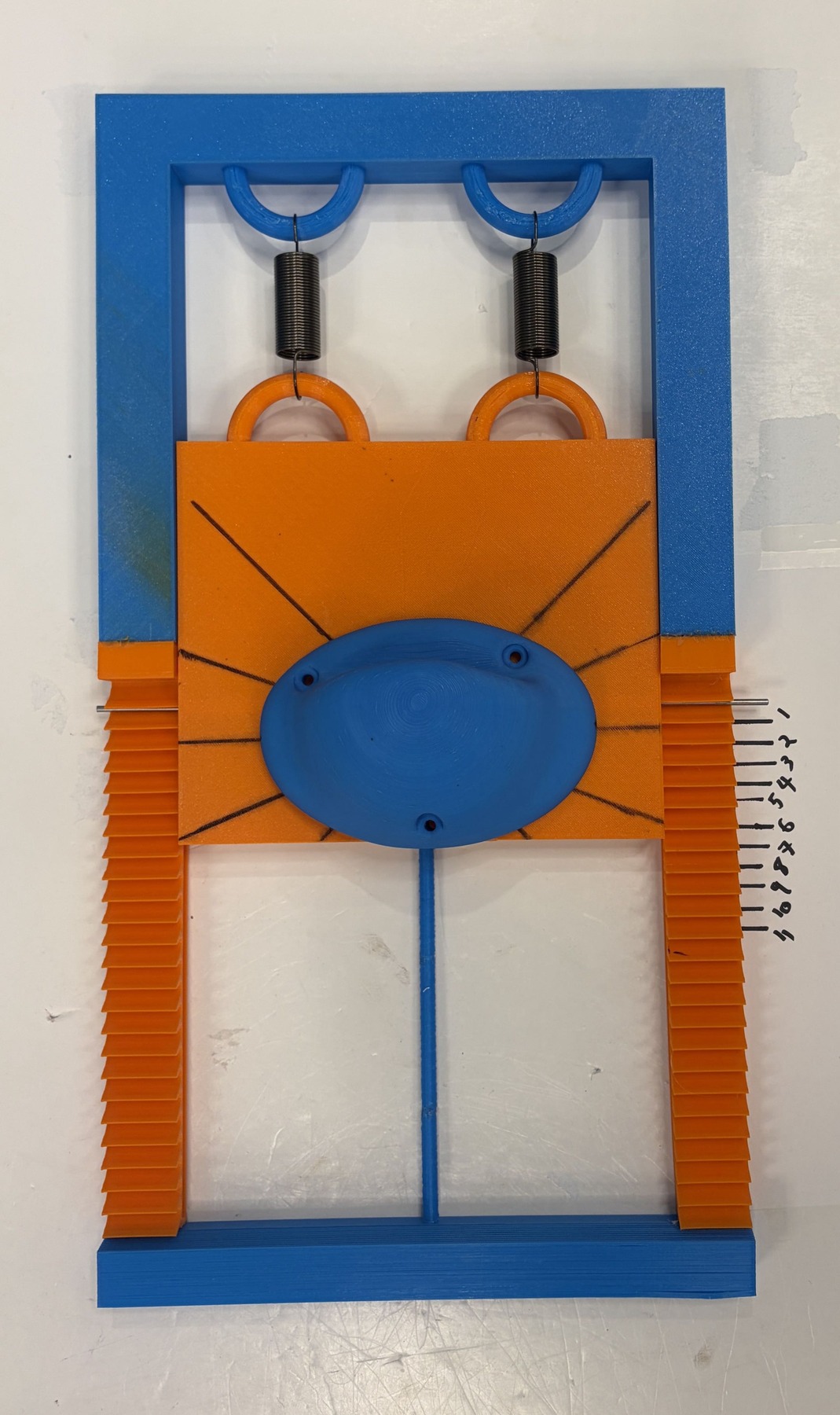} &
\begin{tabular}[b]{@{}cc@{}}
\includegraphics[height=2.05cm]{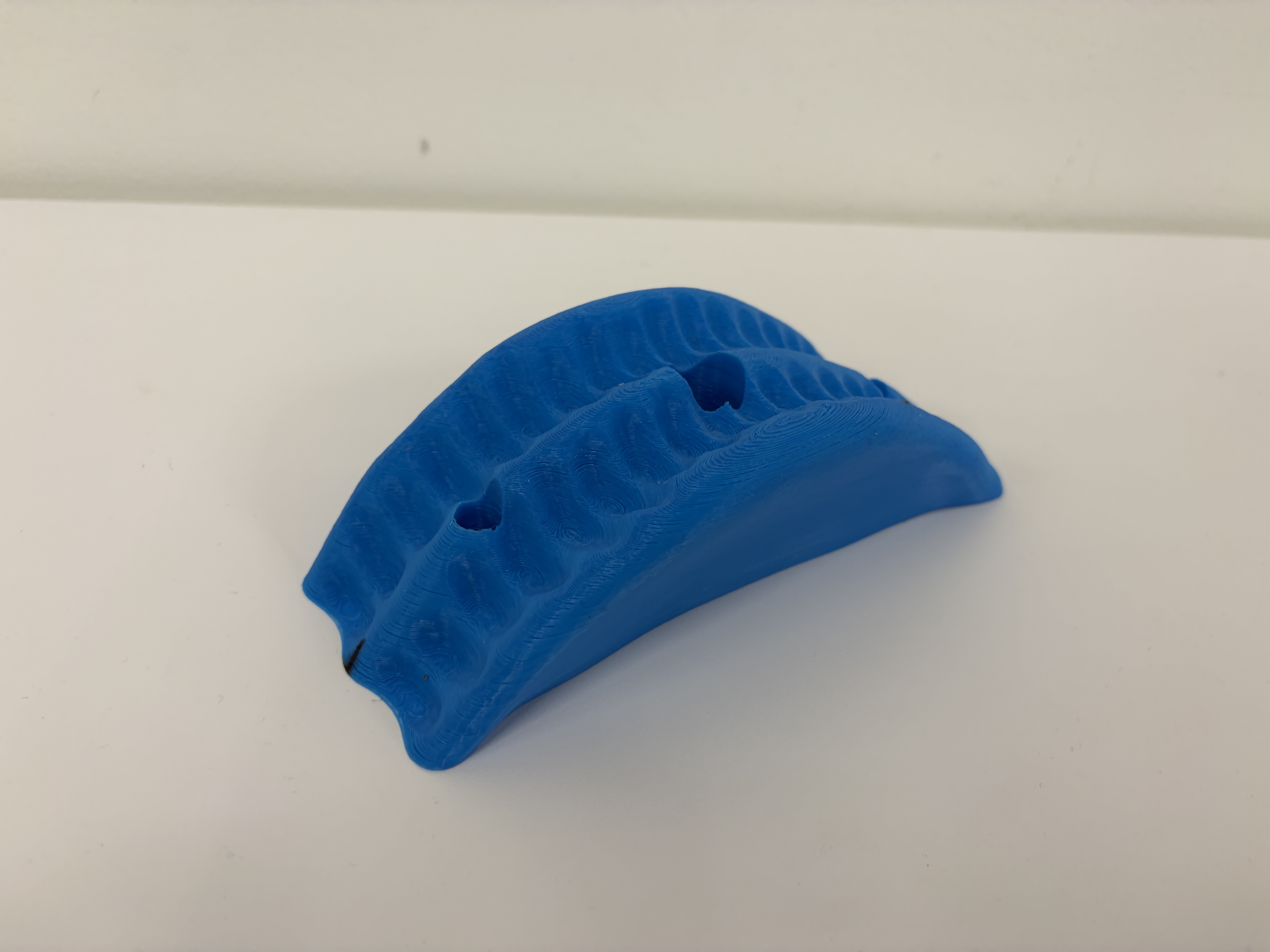} &
\includegraphics[height=2.05cm]{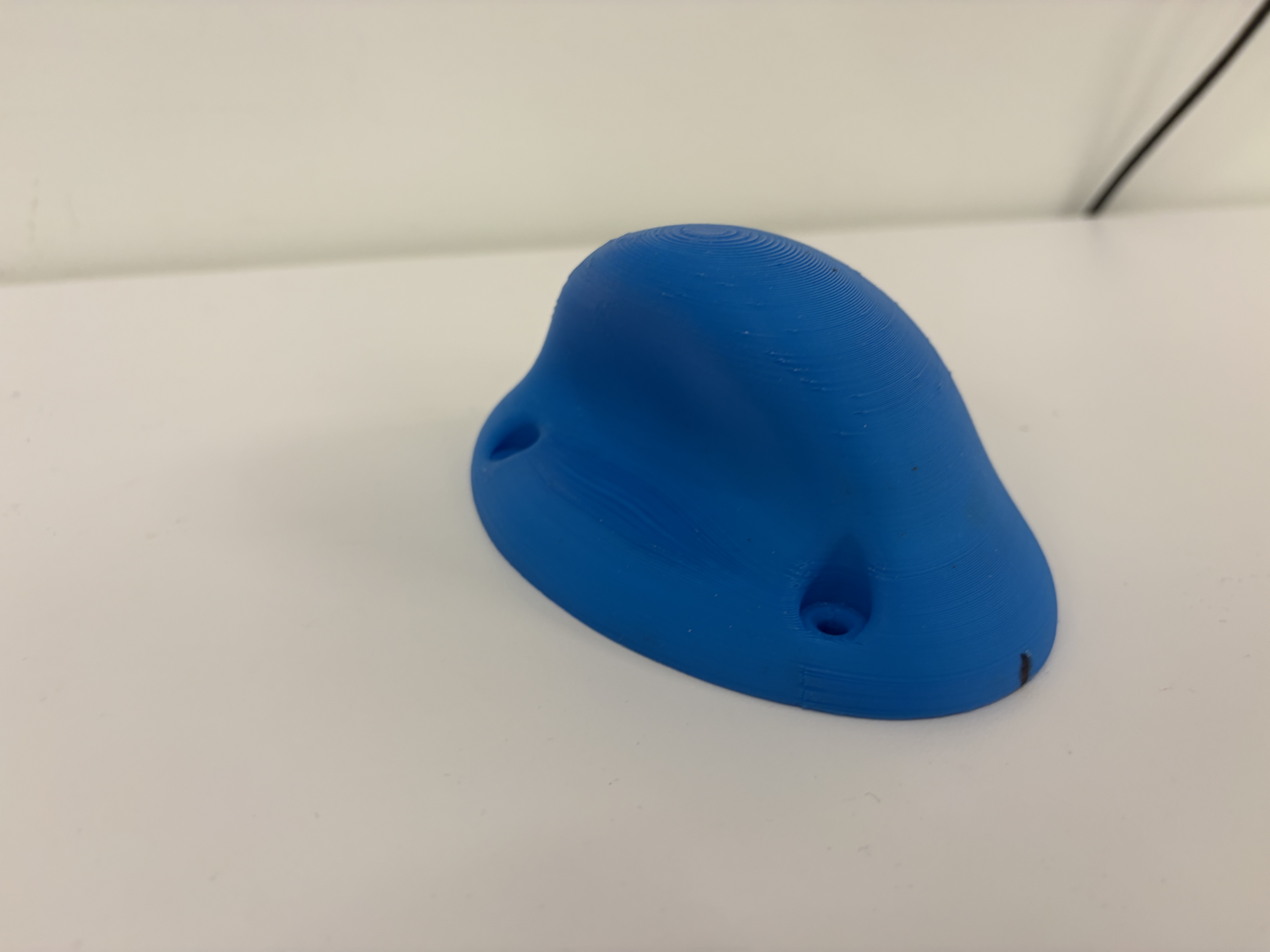} \\
\includegraphics[height=2.05cm]{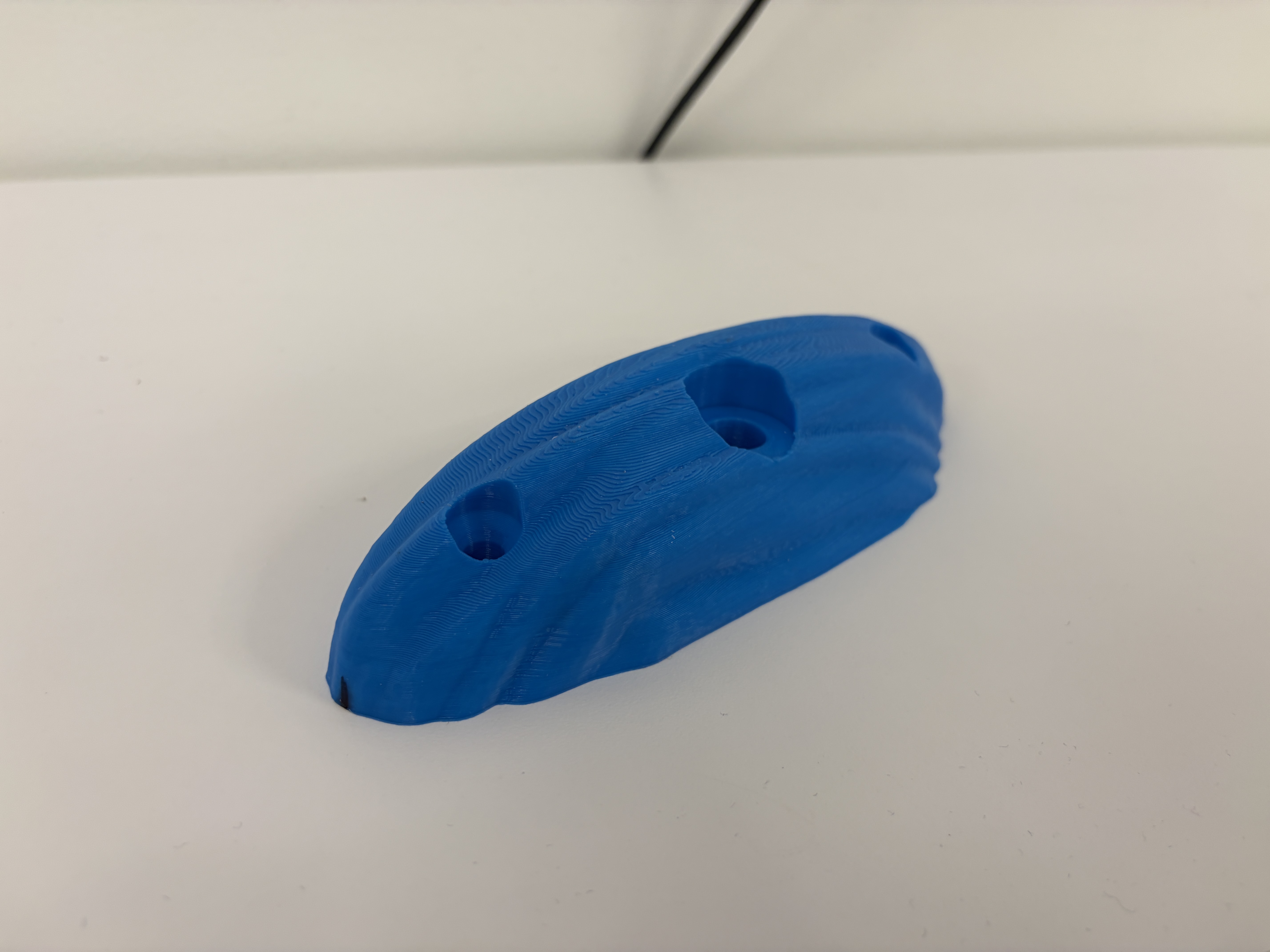} &
\includegraphics[height=2.05cm]{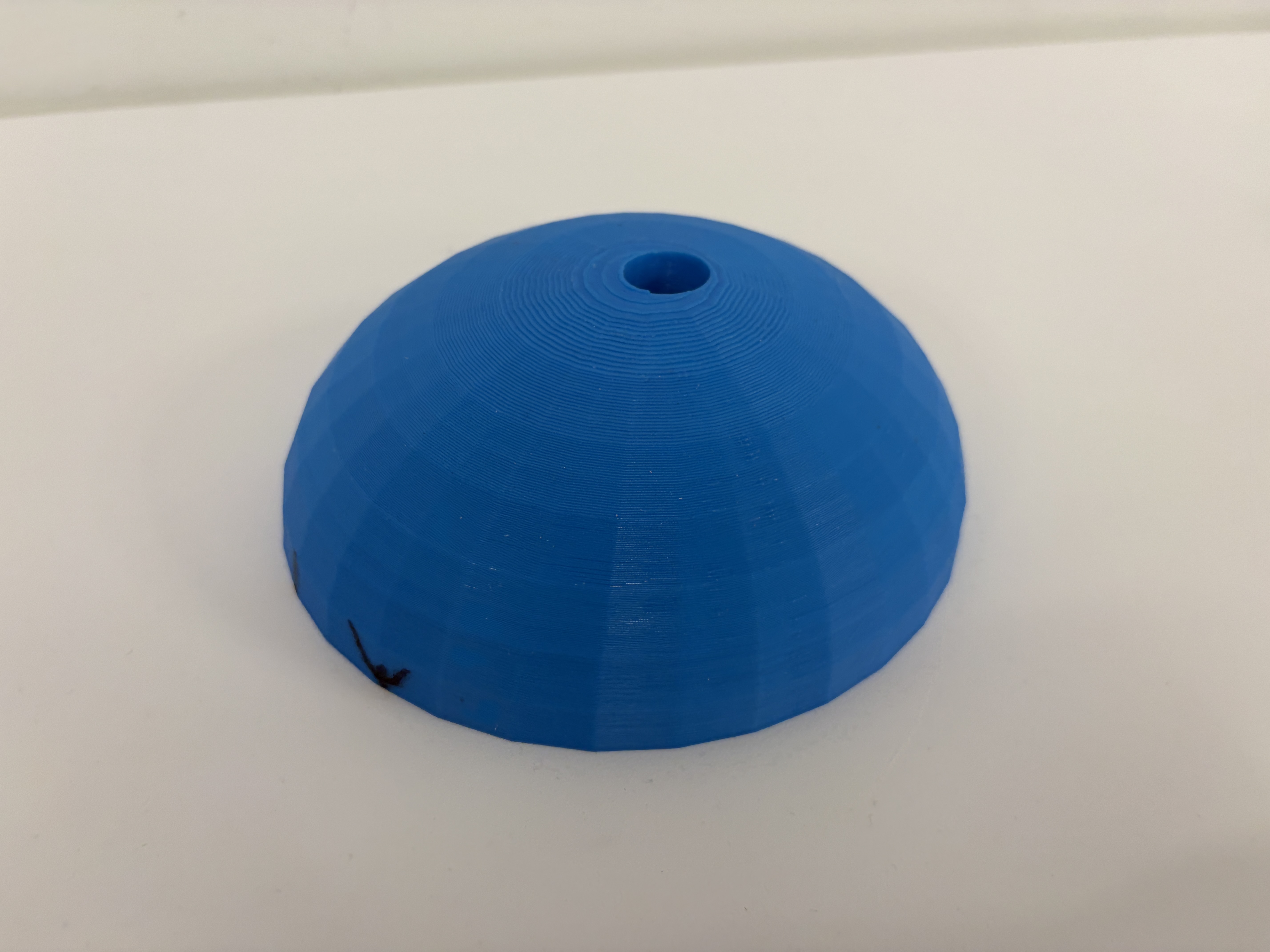}
\end{tabular} \\
\small (a) & \small (b) & \small (c)
\end{tabular}
\caption{Spring-testbed benchmark. (a) The LEAP Hand v1 (16-DoF, on a Franka Emika
Panda) executes a taxonomy-conditioned grasp on a 3D-printed jug hold seated on the
testbed. (b) Top-down view: two extension springs in parallel (top) pull the orange
deck back along a single axis directed toward the robot base, a center guide rod keeps
the deck on that axis, and a pin through the deck rides over the toothed ratchet strips
on both sides, holding the deck at its farthest point; the marked teeth (1--11) show
how far the hold traveled before the grasp slipped. Lines drawn on the deck guide the
five hold orientations. (c) The four
3D-printed climbing holds: pinch (narrow elongated protrusion), jug (domed
body with deep finger pocket), crimp (thin edge with shallow finger ridges), and sloper
(convex hemispherical surface)}
\label{fig:setup}
\end{figure*}

\subsection{Architecture}
\label{sec:arch}
Our policy follows the DP3 design \cite{ze20243d}: a point-cloud encoder
produces a 3D scene embedding, fused with robot state and a grasp-type embedding to
condition a 1D temporal U-Net that parameterizes the diffusion denoiser
(Fig.~\ref{fig:arch}).

\paragraph*{Point cloud encoder}
A PointNet-style encoder \cite{qi2017pointnet} (channels $3 \to 64 \to 128 \to 256$,
no T-Net, max-pool, linear projection) maps $P \in \mathbb{R}^{1024 \times 3}$ to a
256-dimensional feature vector $z_P$. We use XYZ-only points (no color) to reduce
sensitivity to appearance and lighting \cite{ze20243d}.

\paragraph*{State and grasp-type conditioning}
Robot proprioception over $T_o = 2$ timesteps is embedded via an MLP to obtain
$z_s \in \mathbb{R}^{128}$. The grasp-type label $g$ is one-hot encoded and passed
through a small MLP to obtain $z_g \in \mathbb{R}^{64}$. The conditioning vector is
$z = \mathrm{MLP}([z_P;\, z_s;\, z_g]) \in \mathbb{R}^{512}$.
The baseline is the same network without the $z_g$ branch, trained on the same data. We
call it \emph{DP3} for short, but it is our own implementation of the DP3 design
\cite{ze20243d}, not the released DP3 code. The encoder, state MLP, diffusion backbone,
and all hyperparameters are identical between the two.

\paragraph*{Diffusion model}
The denoiser is a 1D temporal U-Net \cite{ronneberger2015unet} with down\_dims
$(256, 512, 1024)$ and feature-wise conditioning on $z$ \cite{perez2018film}, trained
with a cosine $\beta$ schedule \cite{nichol2021iddpm} and 100 DDPM
steps. At inference we use DDIM sampling with 10 denoising steps for 10\,Hz real-time
control \cite{song2021ddim}.

\subsection{Training}
\label{sec:training}
Training follows standard DDPM supervision \cite{ho2020ddpm}: sample diffusion step
$k$, add noise to the ground-truth action chunk $a_0$, and minimize the noise-prediction
loss conditioned on observations and (if applicable) grasp type. We use AdamW
\cite{loshchilov2019adamw} with learning rate $10^{-4}$, $\beta=(0.95, 0.999)$, and
weight decay $10^{-6}$; batch size 128; 3000 epochs with 500-step linear warmup and
cosine decay; power-law EMA with max decay 0.9999; gradient clip 1.0; and PyTorch AMP
mixed precision. Point clouds are augmented with Gaussian jitter ($\sigma = 0.002$\,m)
and 5\% random point dropout. State and action are min--max normalized to $[-1, 1]$.
% -----------------------------------------------------------------------
\section{Spring-Testbed Benchmark}
\label{sec:benchmark}
% -----------------------------------------------------------------------

\subsection{Robot Platform and Sensing}
The benchmark uses a Franka Emika Panda arm (7-DoF) paired with a LEAP Hand v1 \cite{shaw2023leap}
(16-DoF, 4 fingers $\times$ 4 joints each). Arm control is
provided by frankapy \cite{zhang2020frankapy} under Cartesian impedance control. Scene
observations are fused point clouds from four Intel RealSense D415/D455 cameras
(848$\times$480, 30\,Hz), transformed to a common world frame via offline checkerboard
extrinsics calibration, cropped to a workspace bounding box, outlier-filtered, and
downsampled to exactly 1024 points via farthest-point sampling. The state and action
spaces are both 23-dimensional (7 arm joints + 16 hand joints, absolute radians).
The control loop runs at 10\,Hz with per-step joint deltas clamped to
$\pm 0.05$\,rad for safety.

\subsection{Spring Displacement Testbed and Ratcheting Mechanism}
\label{sec:testbed}
Evaluation uses a linearly constrained spring platform that seats each climbing hold
(3D-printed from Thingiverse models
\cite{hold_crimp,hold_sloper,hold_pinch,hold_jug}) and restricts displacement to a
single axis directed toward the robot base
($180^\circ$ in the world $XY$-plane). Two identical extension springs in parallel
provide the restoring force, empirically calibrated as
\[
F(x) = 5.25\,\text{N} + 0.28\,\text{N/mm} \times x,
\]
where $x$ is the hold's displacement in millimeters.
A linear pawl ratchet (11 teeth, 9.3\,mm per tooth, 102.3\,mm maximum travel) captures
the hold's peak displacement even when the hold slips mid-trial
(Fig.~\ref{fig:setup}a,b). After each rollout the operator reads the ratchet tooth count
$n \in \{0, \dots, 11\}$, from which the script computes displacement and slip force.
Each trial therefore gives a slip force in newtons, in steps of about 2.6\,N, instead of
a pass/fail outcome. Holds are evaluated at five orientations relative to the pull axis:
$0^\circ$, $\pm 22.5^\circ$, and $\pm 45^\circ$. We did not test beyond $\pm 45^\circ$,
where we expected the hold to slip regardless of the grasp.

\subsection{Dataset}
\label{sec:dataset}
The training set contains 200 teleoperated demonstrations (50 per grasp type $\times$
4 types) totaling 27{,}821 timesteps at 10\,Hz (46.4 minutes of teleop). Episode
lengths range from 61 to 279 steps (median 132). All episodes were collected by a
single operator via Meta Quest 2 VR hand-tracking, on the same spring testbed used for
evaluation. For each hold, the operator recorded 10 episodes at one orientation before
rotating the hold to the next, giving 10 episodes at each of the five orientations (50
per hold, 200 in total). Orientation is not given to the policy. Every episode starts
with the same reset posture.
Each episode stores time-aligned 23-dim state and 23-dim joint-space action, the
grasp-type label, and one fused point cloud captured with the arm parked out of view of
all four cameras. The hold does not move during a demonstration, so one scan per episode
is enough.

\subsection{Paired Evaluation Protocol}
\label{sec:protocol}
We evaluate the two policies in pairs. Both trials of a pair use the same physical hold
position, and the order alternates from pair to pair, so each model runs first in
exactly 10 of the 20 pairs for every grasp type. We capture a fresh point cloud before
every trial, because the first trial's contact can nudge the hold (median shift
1.2\,mm between the two scans of a pair). After each rollout the arm pulls along the
testbed's axis under Cartesian impedance control, with stiffness $k_x = 4000$\,N/m along
the pull, $k_y = 100$\,N/m laterally, and $k_z = 2000$\,N/m vertically to hold up the
$\sim$1\,kg LEAP hand. The pull is 13\,cm. The operator then enters the ratchet reading.

% -----------------------------------------------------------------------
\section{Experiments}
\label{sec:exp}
% -----------------------------------------------------------------------
Prior to the paired evaluation we stated two hypotheses: \textbf{H1}
(primary): BetaGrasp achieves higher slip force than DP3 on all four grasp types; and
\textbf{H2} (ordering): the gains are largest on the grasps that need precise finger
placement (crimp, sloper, pinch) and smallest on the jug, which tolerates imprecise
placement.

\subsection{Mechanistic Analysis}
\label{sec:mechanistic}
One explanation for why the label would help is \emph{mode selection}: without it, a
policy trained on several grasp types may fall back on a single behavior instead of the
grasp each hold needs. We test this with two offline analyses of the trained
policies.\footnote{The latent analysis uses the pre-rig checkpoints and dataset,
collected on the flat table before the spring testbed was built. The
closed-loop rollouts use the spring-testbed checkpoints evaluated on hardware in
\S\ref{sec:robot_eval}.}

\begin{figure}[t]
\centering
\includegraphics[width=\columnwidth]{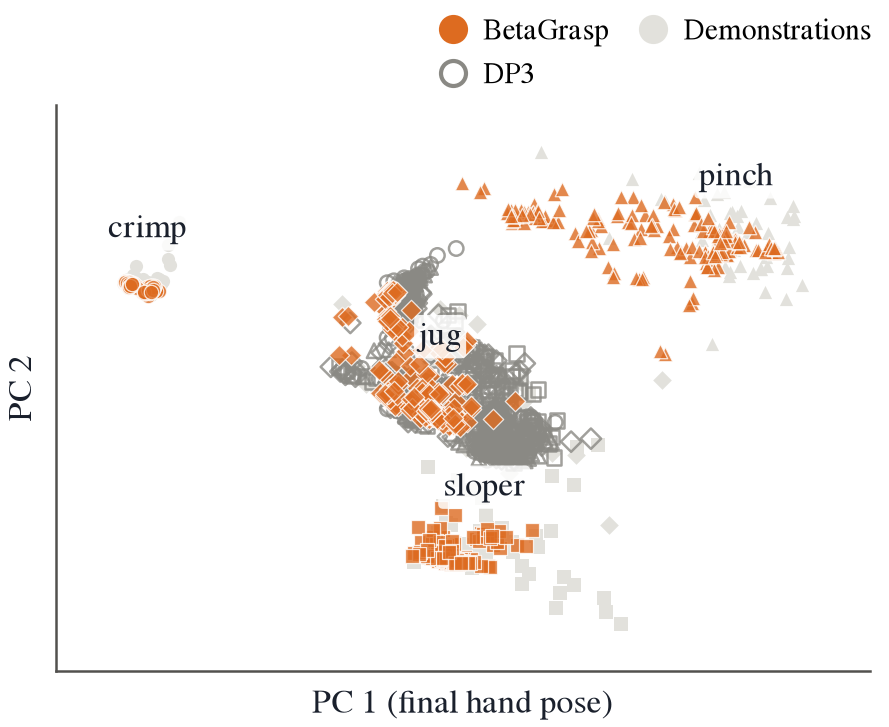}
\caption{Final hand poses of simulated closed-loop rollouts started from the deployment
reset pose, projected onto the first two principal components of the demonstrators' final
hand poses (light gray; marker shape = hold). BetaGrasp rollouts (orange) mostly land
on the demonstrated grasp for their hold; DP3 rollouts (gray outline) end in the jug grasp on every hold.}
\label{fig:mech}
\end{figure}

\paragraph*{Latent-space separation}
We pass the first frame of every pre-rig training episode through DP3's encoder and
measure how well the grasp types separate, using the silhouette score and 5-NN accuracy
under stratified 5-fold cross-validation. With no label, DP3's fused encoder output
reaches 99\% 5-NN accuracy and silhouette 0.32. The point-cloud branch alone does not
separate the types (silhouette $-0.03$, 5-NN accuracy 39\%, chance 25\%), so the
separation comes from the proprioception channel.

\paragraph*{Closed-loop rollouts from the reset pose}
On hardware, every trial starts from one fixed reset pose. We simulate each policy in
closed loop from that reset pose, for 25 inputs per grasp type and $K=8$
rollouts each (commanded joints tracked exactly), and assign each final
hand pose to the nearest demonstrator grasp-type mean. BetaGrasp reaches the correct grasp in 94\% (crimp, sloper, jug 100\%;
pinch 76\%, with the misses landing on jug), and DP3 in only 25\%: all 800 of its
rollouts end in a jug grasp, whatever the hold (Fig.~\ref{fig:mech}).

\paragraph*{Interpretation}
These results fit the mode-selector account. In our analysis the point-cloud encoder
does not separate the four holds, so at deployment, where every trial starts from the
same reset pose, nothing in DP3's input identifies the grasp and it falls back on a
single one. The label supplies the missing information.

\subsection{Hardware Evaluation}
\label{sec:robot_eval}
We ran 80 paired trials, or 160 rollouts (4 pairs per orientation $\times$ 5
orientations $\times$ 4 holds), in a single $\sim$7-hour session with a single
operator.

\begin{figure}[t]
\centering
\includegraphics[width=\columnwidth]{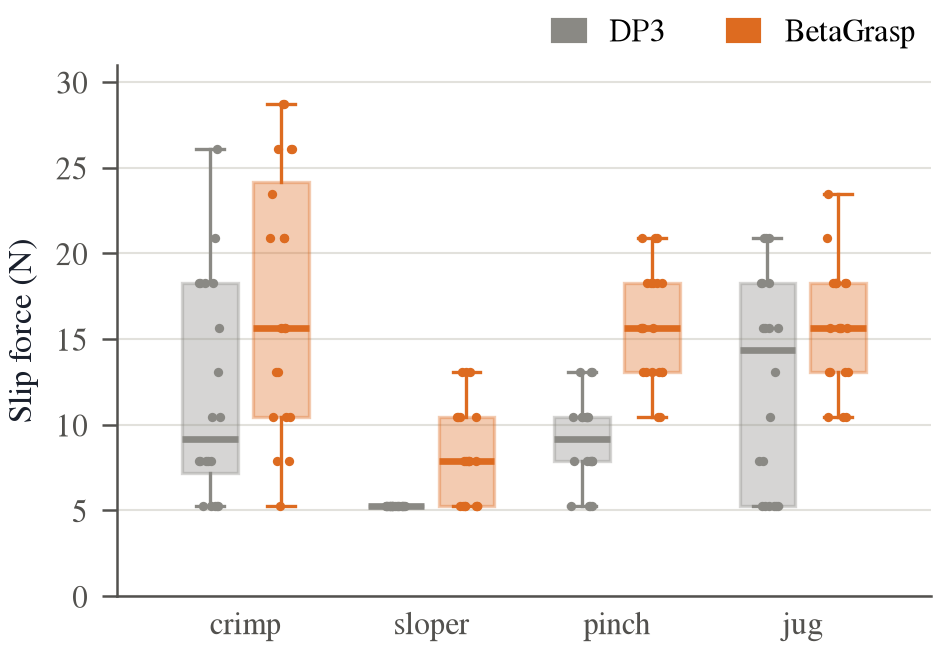}
\caption{Slip force per hold on the spring testbed, 20 trials per policy per hold.
Boxes show the interquartile range and median; dots are individual trials. The
5.2\,N floor is the spring preload, reached when the grasp slips off before the hold
moves.}
\label{fig:hw}
\end{figure}

\begin{table}[t]
\centering\small
\setlength{\tabcolsep}{4pt}
\caption{Slip force per hold, averaged over 20 trials per policy, and the share of
trials where the hand slipped off before moving the hold (5.2\,N preload).}
\label{tab:hw}
\begin{tabular}{lcccc}
\toprule
& \multicolumn{2}{c}{Avg.\ slip force (N)} & \multicolumn{2}{c}{Slipped off (\%)} \\
\cmidrule(lr){2-3}\cmidrule(lr){4-5}
Hold & DP3 & BetaGrasp & DP3 & BetaGrasp \\
\midrule
Crimp  & 11.8 & 17.4 & 25  & 5  \\
Sloper & 5.2  & 8.2  & 100 & 35 \\
Pinch  & 9.0  & 15.8 & 20  & 0  \\
Jug    & 12.5 & 15.8 & 30  & 0  \\
\midrule
All    & 9.6  & 14.3 & 44  & 10 \\
\bottomrule
\end{tabular}
\end{table}

\paragraph*{H1: slip force}
BetaGrasp's average slip force is higher on all four holds (Fig.~\ref{fig:hw},
Table~\ref{tab:hw}). Overall it rises from 9.6\,N to 14.3\,N ($+4.7$\,N), and across the
80 pairs BetaGrasp beats DP3 in 61, ties 14, and loses 5.

\paragraph*{H2: ordering across grasp types}
H2 holds only in part. The gains are largest on pinch ($+6.8$\,N, with BetaGrasp
winning all 20 pairs) and crimp ($+5.6$\,N). The jug gain is smaller ($+3.3$\,N), and
the jug is DP3's strongest hold (12.5\,N). Jug is the grasp DP3 fell back on in our
simulated rollouts, so on the jug hold it likely performed the right grasp; we did not
log which grasp was executed on hardware. The sloper gain is the smallest ($+3.0$\,N),
but DP3's sloper average sits at the 5.2\,N floor: it slipped off in all 20 trials.

\paragraph*{Slip-offs}
Much of the gain comes from fewer grasps slipping straight off the hold. DP3 slipped off
before moving the hold in 44\% of trials; BetaGrasp did in 10\% (Table~\ref{tab:hw}).

\paragraph*{Orientation profile}
For BetaGrasp, every grasp type is weakest at $+45^\circ$ (pinch ties at $-45^\circ$).
Crimp varies most with orientation (from 26.8\,N at $-22.5^\circ$ to 7.9\,N at $+45^\circ$),
consistent with the thin crimp edge leaving little angular tolerance. With only four
trials per orientation, these are trends rather than firm differences.

% -----------------------------------------------------------------------
\section{Limitations and Future Work}
\label{sec:limitations}
% -----------------------------------------------------------------------

BetaGrasp was tested on a single embodiment, a Franka arm with a LEAP Hand. The label
enters through a small separate branch (0.1\% of the parameters), so we expect the same
conditioning to carry over to other dexterous hands such as the Allegro or Shadow Hand,
but testing it there is the most direct next step. The training data come from one
operator and one physical hold per grasp type, so part of each per-type gain may come
from that particular hold rather than the grasp type; demonstrations from several
operators on several holds per type, evaluated on holds held out from training, would
separate the two. The grasp label is currently supplied by the operator, and a
vision-language model that predicts it from an image of the hold, as OmniDexVLG
\cite{omnidexvlg2024} does for grasp poses, would make the system fully autonomous.

The spring testbed loads the grasp along one fixed axis; a wall-mounted or multi-axis
rig would test the multi-directional and sustained loads of real climbing. The
closed-loop rollouts in \S\ref{sec:mechanistic} are simulated (commanded joints tracked
exactly) and the latent analysis uses the pre-rig checkpoints, so repeating
both on the spring-testbed models, with rollouts on hardware, would confirm them. Finally, BetaGrasp's 14.3\,N average is
a moderate grip, which more demonstrations per grasp type or per-grasp tuning may
improve.

% -----------------------------------------------------------------------
\section{Conclusion}
% -----------------------------------------------------------------------
We presented BetaGrasp, a grasp-taxonomy-conditioned 3D diffusion policy for grasping
climbing holds with a dexterous hand, and a climbing-hold benchmark built on a spring
testbed whose ratchet records slip force. Started from its reset pose without the grasp
label, our policy fell back on a single grasp in simulation; with the label, it held on
with more force on every hold on hardware (14.3\,N versus 9.6\,N on average). The label
acts as a switch for which grasp the policy performs.

\paragraph*{Reproducibility}
Code, data, checkpoints, and the CAD files, 3D-print files, and parts list for the
spring testbed will be released.

\bibliographystyle{IEEEtran}
\bibliography{references}

@article{chi2025diffusion,
  title={Diffusion policy: Visuomotor policy learning via action diffusion},
  author={Chi, Cheng and Xu, Zhenjia and Feng, Siyuan and Cousineau, Eric and Du, Yilun and Burchfiel, Benjamin and Tedrake, Russ and Song, Shuran},
  journal={The International Journal of Robotics Research},
  volume={44},
  number={10-11},
  pages={1684--1704},
  year={2025},
  publisher={Sage Publications, Sage UK: London, England}
}

@article{ze20243d,
  title={3d diffusion policy: Generalizable visuomotor policy learning via simple 3d representations},
  author={Ze, Yanjie and Zhang, Gu and Zhang, Kangning and Hu, Chenyuan and Wang, Muhan and Xu, Huazhe},
  journal={arXiv preprint arXiv:2403.03954},
  year={2024}
}

@inproceedings{ze2024idp3,
  title     = {Generalizable Humanoid Manipulation with Improved 3D Diffusion Policies},
  author    = {Ze, Yanjie and Chen, Zixuan and Wang, Wenhao and Chen, Tianyi and
               He, Xialin and Yuan, Ying and Peng, Xue Bin and Wu, Jiajun},
  booktitle = {Conference on Robot Learning (CoRL)},
  year      = {2024},
  url       = {https://arxiv.org/abs/2410.10803}
}

@inproceedings{ho2020ddpm,
  title     = {Denoising Diffusion Probabilistic Models},
  author    = {Ho, Jonathan and Jain, Ajay and Abbeel, Pieter},
  booktitle = {Advances in Neural Information Processing Systems (NeurIPS)},
  year      = {2020},
  url       = {https://arxiv.org/abs/2006.11239}
}

@inproceedings{song2021ddim,
  title     = {Denoising Diffusion Implicit Models},
  author    = {Song, Jiaming and Meng, Chenlin and Ermon, Stefano},
  booktitle = {International Conference on Learning Representations (ICLR)},
  year      = {2021},
  url       = {https://arxiv.org/abs/2010.02502}
}

@inproceedings{qi2017pointnet,
  title     = {{PointNet}: Deep Learning on Point Sets for 3D Classification
               and Segmentation},
  author    = {Qi, Charles R. and Su, Hao and Mo, Kaichun and Guibas, Leonidas J.},
  booktitle = {Proceedings of the IEEE Conference on Computer Vision and
               Pattern Recognition (CVPR)},
  year      = {2017},
  url       = {https://arxiv.org/abs/1612.00593}
}

@inproceedings{wang2024gendp,
  title     = {{GenDP}: 3D Semantic Fields for Category-Level Generalizable
               Diffusion Policy},
  author    = {Wang, Yixuan and Yin, Guang and Huang, Binghao and Kelestemur, Tarik
               and Wang, Jiuguang and Li, Yunzhu},
  booktitle = {Conference on Robot Learning (CoRL)},
  year      = {2024},
  url       = {https://arxiv.org/abs/2410.17488}
}

@inproceedings{wang2024dexcap,
  title     = {{DexCap}: Scalable and Portable Mocap Data Collection System for
               Dexterous Manipulation},
  author    = {Wang, Chen and Shi, Haochen and Wang, Weizhuo and Zhang, Ruohan and
               Fei-Fei, Li and Liu, C. Karen},
  booktitle = {Robotics: Science and Systems (RSS)},
  year      = {2024},
  url       = {https://arxiv.org/abs/2403.07788}
}

@article{george2023openvr,
  title     = {{OpenVR}: Teleoperation for Manipulation},
  author    = {George, Abraham and Bartsch, Alison and Farimani, Amir Barati},
  journal   = {arXiv preprint arXiv:2305.09765},
  year      = {2023},
  url       = {https://arxiv.org/abs/2305.09765}
}

@inproceedings{wu2024unidexfpm,
  title     = {{UniDexFPM}: Universal Dexterous Functional Pre-Grasp Manipulation
               via Diffusion Policy},
  author    = {Wu, Tianhao and Gan, Yunchong and Wu, Mingdong and Cheng, Jingbo and
               Yang, Yaodong and Zhu, Yixin and Dong, Hao},
  booktitle = {IEEE International Conference on Robotics and Automation (ICRA)},
  year      = {2024},
  url       = {https://arxiv.org/abs/2403.12421}
}

@article{weng2024dexdiffuser,
  title   = {{DexDiffuser}: Generating Dexterous Grasps with Diffusion Models},
  author  = {Weng, Zehang and Lu, Haofei and Kragic, Danica and Lundell, Jens},
  journal = {IEEE Robotics and Automation Letters},
  year    = {2024},
  url     = {https://arxiv.org/abs/2402.02989}
}

@inproceedings{crossdex2025,
  title     = {Cross-Embodiment Dexterous Grasping with Reinforcement Learning},
  author    = {Yuan, Haoqi and Zhou, Bohan and Fu, Yuhui and Lu, Zongqing},
  booktitle = {International Conference on Learning Representations (ICLR)},
  year      = {2025},
  url       = {https://arxiv.org/abs/2410.02479}
}

@article{feix2016grasp,
  title   = {The {GRASP} Taxonomy of Human Grasp Types},
  author  = {Feix, Thomas and Romero, Javier and Schmiedmayer, Heinz-Bodo and
             Dollar, Aaron M. and Kragic, Danica},
  journal = {IEEE Transactions on Human-Machine Systems},
  volume  = {46},
  number  = {1},
  pages   = {66--77},
  year    = {2016},
  doi     = {10.1109/THMS.2015.2470657}
}

@inproceedings{dexonomy2025,
  title     = {Dexonomy: Synthesizing All Dexterous Grasp Types in a Grasp Taxonomy},
  author    = {Chen, Jiayi and Ke, Yubin and Peng, Lin and Wang, He},
  booktitle = {Robotics: Science and Systems (RSS)},
  year      = {2025},
  url       = {https://arxiv.org/abs/2504.18829}
}

@inproceedings{tian2024graspasyousay,
  title     = {Grasp as You Say: Language-guided Dexterous Grasp Generation},
  author    = {Wei, Yi-Lin and Jiang, Jian-Jian and Xing, Chengyi and Tan, Xian-Tuo
               and Wu, Xiao-Ming and Li, Hao and Cutkosky, Mark and Zheng, Wei-Shi},
  booktitle = {Advances in Neural Information Processing Systems (NeurIPS)},
  year      = {2024},
  url       = {https://arxiv.org/abs/2405.19291}
}

@article{omnidexvlg2024,
  title   = {{OmniDexVLG}: Learning Dexterous Grasp Generation from Vision Language
             Model-Guided Grasp Semantics, Taxonomy and Functional Affordance},
  author  = {Zhang, Lei and Zheng, Diwen and Bai, Kaixin and Bing, Zhenshan and
             Marton, Zoltan-Csaba and Chen, Zhaopeng and Knoll, Alois Christian and
             Zhang, Jianwei},
  journal = {arXiv preprint arXiv:2512.03874},
  year    = {2025},
  url     = {https://arxiv.org/abs/2512.03874}
}

@inproceedings{dexgraspvla2026,
  title     = {{DexGraspVLA}: A Vision-Language-Action Framework Towards General
               Dexterous Grasping},
  author    = {Zhong, Yifan and Huang, Xuchuan and Li, Ruochong and Zhang, Ceyao and
               Chen, Zhang and Guan, Tianrui and Zeng, Fanlian and Lui, Ka Num and
               Ye, Yuyao and Liang, Yitao and Yang, Yaodong and Chen, Yuanpei},
  booktitle = {AAAI Conference on Artificial Intelligence},
  year      = {2026},
  url       = {https://arxiv.org/abs/2502.20900}
}

@inproceedings{loshchilov2019adamw,
  title     = {Decoupled Weight Decay Regularization},
  author    = {Loshchilov, Ilya and Hutter, Frank},
  booktitle = {International Conference on Learning Representations (ICLR)},
  year      = {2019},
  url       = {https://openreview.net/forum?id=Bkg6RiCqY7}
}

@inproceedings{shaw2023leap,
  title     = {{LEAP Hand}: Low-Cost, Efficient, and Anthropomorphic Hand for
               Robot Learning},
  author    = {Shaw, Kenneth and Agarwal, Ananye and Pathak, Deepak},
  booktitle = {Robotics: Science and Systems (RSS)},
  year      = {2023},
  url       = {https://arxiv.org/abs/2309.06440}
}

@article{zhang2020frankapy,
  title   = {A Modular Robotic Arm Control Stack for Research: {Franka-Interface}
             and {FrankaPy}},
  author  = {Zhang, Kevin and Sharma, Mohit and Liang, Jacky and Kroemer, Oliver},
  journal = {arXiv preprint arXiv:2011.02398},
  year    = {2020},
  url     = {https://arxiv.org/abs/2011.02398}
}

@inproceedings{perez2018film,
  title     = {{FiLM}: Visual Reasoning with a General Conditioning Layer},
  author    = {Perez, Ethan and Strub, Florian and de Vries, Harm and Dumoulin,
               Vincent and Courville, Aaron},
  booktitle = {AAAI Conference on Artificial Intelligence},
  year      = {2018},
  url       = {https://arxiv.org/abs/1709.07871}
}

@inproceedings{ronneberger2015unet,
  title     = {{U-Net}: Convolutional Networks for Biomedical Image Segmentation},
  author    = {Ronneberger, Olaf and Fischer, Philipp and Brox, Thomas},
  booktitle = {Medical Image Computing and Computer-Assisted Intervention (MICCAI)},
  year      = {2015},
  url       = {https://arxiv.org/abs/1505.04597}
}

@inproceedings{nichol2021iddpm,
  title     = {Improved Denoising Diffusion Probabilistic Models},
  author    = {Nichol, Alexander Quinn and Dhariwal, Prafulla},
  booktitle = {International Conference on Machine Learning (ICML)},
  year      = {2021},
  url       = {https://arxiv.org/abs/2102.09672}
}

@inproceedings{zhao2023aloha,
  title     = {Learning Fine-Grained Bimanual Manipulation with Low-Cost Hardware},
  author    = {Zhao, Tony Z. and Kumar, Vikash and Levine, Sergey and Finn, Chelsea},
  booktitle = {Robotics: Science and Systems (RSS)},
  year      = {2023},
  url       = {https://arxiv.org/abs/2304.13705}
}

@inproceedings{qin2023anyteleop,
  title     = {{AnyTeleop}: A General Vision-Based Dexterous Robot Arm-Hand
               Teleoperation System},
  author    = {Qin, Yuzhe and Yang, Wei and Huang, Binghao and Van Wyk, Karl and
               Su, Hao and Wang, Xiaolong and Chao, Yu-Wei and Fox, Dieter},
  booktitle = {Robotics: Science and Systems (RSS)},
  year      = {2023},
  url       = {https://arxiv.org/abs/2307.04577}
}

@inproceedings{cheng2024opentelevision,
  title     = {{Open-TeleVision}: Teleoperation with Immersive Active Visual Feedback},
  author    = {Cheng, Xuxin and Li, Jialong and Yang, Shiqi and Yang, Ge and
               Wang, Xiaolong},
  booktitle = {Conference on Robot Learning (CoRL)},
  year      = {2024},
  url       = {https://arxiv.org/abs/2407.01512}
}

@inproceedings{wang2023dexgraspnet,
  title     = {{DexGraspNet}: A Large-Scale Robotic Dexterous Grasp Dataset for
               General Objects Based on Simulation},
  author    = {Wang, Ruicheng and Zhang, Jialiang and Chen, Jiayi and Xu, Yinzhen and
               Li, Puhao and Liu, Tengyu and Wang, He},
  booktitle = {IEEE International Conference on Robotics and Automation (ICRA)},
  year      = {2023},
  url       = {https://arxiv.org/abs/2210.02697}
}

@inproceedings{xu2023unidexgrasp,
  title     = {{UniDexGrasp}: Universal Robotic Dexterous Grasping via Learning
               Diverse Proposal Generation and Goal-Conditioned Policy},
  author    = {Xu, Yinzhen and Wan, Weikang and Zhang, Jialiang and Liu, Haoran and
               Shan, Zikang and Shen, Hao and Wang, Ruicheng and Geng, Haoran and
               Weng, Yijia and Chen, Jiayi and Liu, Tengyu and Yi, Li and Wang, He},
  booktitle = {Proceedings of the IEEE/CVF Conference on Computer Vision and
               Pattern Recognition (CVPR)},
  year      = {2023},
  url       = {https://arxiv.org/abs/2303.00938}
}

@article{napier1956prehensile,
  title   = {The Prehensile Movements of the Human Hand},
  author  = {Napier, John R.},
  journal = {The Journal of Bone and Joint Surgery. British Volume},
  volume  = {38-B},
  number  = {4},
  pages   = {902--913},
  year    = {1956}
}

@article{cutkosky1989grasp,
  title   = {On Grasp Choice, Grasp Models, and the Design of Hands for
             Manufacturing Tasks},
  author  = {Cutkosky, Mark R.},
  journal = {IEEE Transactions on Robotics and Automation},
  volume  = {5},
  number  = {3},
  pages   = {269--279},
  year    = {1989},
  doi     = {10.1109/70.34763}
}

@article{amca2012grip,
  title   = {Effect of Hold Depth and Grip Technique on Maximal Finger Forces in
             Rock Climbing},
  author  = {Amca, Arif Mithat and Vigouroux, Laurent and Aritan, Serdar and
             Berton, Eric},
  journal = {Journal of Sports Sciences},
  volume  = {30},
  number  = {7},
  pages   = {669--677},
  year    = {2012},
  doi     = {10.1080/02640414.2012.658845}
}

@misc{hold_crimp,
  author       = {{BMO\_FUSE}},
  title        = {Not Enough Crimps Sample Pack Climbing Holds},
  howpublished = {Thingiverse, thing:6375816},
  note         = {File \texttt{Sample\_Crimp\_1.stl}},
  url          = {https://www.thingiverse.com/thing:6375816}
}

@misc{hold_sloper,
  author       = {{mjdrum92}},
  title        = {Climbing Hold Sloper 1},
  howpublished = {Thingiverse, thing:2633613},
  url          = {https://www.thingiverse.com/thing:2633613}
}

@misc{hold_pinch,
  author       = {{BMO\_FUSE}},
  title        = {Green Fossilized Incut Pinch Climbing Hold},
  howpublished = {Thingiverse, thing:6320163},
  url          = {https://www.thingiverse.com/thing:6320163}
}

@misc{hold_jug,
  author       = {{SarDou3D}},
  title        = {Jugs Sample Pack},
  howpublished = {Thingiverse, thing:6580253},
  note         = {File \texttt{CH2\_S.stl}},
  url          = {https://www.thingiverse.com/thing:6580253}
}

@article{schweizer2001crimp,
  title   = {Biomechanical Properties of the Crimp Grip Position in Rock Climbers},
  author  = {Schweizer, Andreas},
  journal = {Journal of Biomechanics},
  volume  = {34},
  number  = {2},
  pages   = {217--223},
  year    = {2001},
  doi     = {10.1016/S0021-9290(00)00184-6}
}

\end{document}